%% file: main.tex
\documentclass[10pt,twocolumn,letterpaper]{article}
\usepackage[pagenumbers]{cvpr}
\input{preamble}

\definecolor{cvprblue}{rgb}{0.21,0.49,0.74}
\usepackage[pagebackref,breaklinks,colorlinks,allcolors=cvprblue]{hyperref}
\title{CoT-Factory: Synthesizing the Missing Reasoning in Agent Trajectories}
\author{%
  Tianrui Zhu\textsuperscript{1}\quad
  Ximan Liu\textsuperscript{2}\quad
  Yong Liu\textsuperscript{2,\textdagger}\quad
  Sule Bai\textsuperscript{2}\quad
  Shi-Yu Tian\textsuperscript{2}\\
  Yuji Wang\textsuperscript{1}\quad
  Jingqi Tian\textsuperscript{1}\quad
  Yipeng Sun\textsuperscript{2}\quad
  Simiu Gu\textsuperscript{2}\quad
  Yansong Tang\textsuperscript{1,\textdagger}\\[0.4em]
  \normalsize \textsuperscript{1}Tsinghua Shenzhen International Graduate School\quad
  \textsuperscript{2}JD.com
}
\hypersetup{pdfauthor={Tianrui Zhu, Ximan Liu, Yong Liu, Sule Bai, Shi-Yu Tian,
  Yuji Wang, Jingqi Tian, Yipeng Sun, Simiu Gu, Yansong Tang}}
\begin{document}
\maketitle
\begingroup
  \renewcommand{\thefootnote}{\textdagger}
  \footnotetext[1]{Corresponding authors.}
\endgroup
\suppressfloats[t]

\begin{abstract}
\input{sections/abstract.tex}
\end{abstract}
\input{figures/training-progress-teaser.tex}

\input{sections/introduction.tex}

\input{sections/related-work.tex}

\input{sections/method.tex}

\input{sections/experiments.tex}

\input{sections/conclusion.tex}

\input{sections/ai-use-statement.tex}

{\small
\bibliographystyle{ieeenat_fullname}
\bibliography{references}
}
\end{document}

%% file: preamble.tex
\usepackage[T1]{fontenc}
\usepackage{tabularx}
\usepackage{multirow}
\usepackage{enumitem}
\usepackage{fancyvrb}
\usepackage{colortbl}
\usepackage{afterpage}
\usepackage{algorithm}
\usepackage{algpseudocode}
\newcommand{\scorechange}[2]{%
  \makebox[2.2em][r]{#1}\hspace{0.25em}%
  \makebox[3.1em][l]{{\scriptsize\sffamily\textcolor{black!65}{%
    \if\relax\detokenize{#2}\relax\else(#2)\fi}}}}

\makeatletter
\newsavebox\pandoc@box
\newcommand*\pandocbounded[1]{%
  \sbox\pandoc@box{#1}%
  \Gscale@div\@tempa{\textheight}{\dimexpr\ht\pandoc@box+\dp\pandoc@box\relax}%
  \Gscale@div\@tempb{\linewidth}{\wd\pandoc@box}%
  \ifdim\@tempb\p@<\@tempa\p@\let\@tempa\@tempb\fi%
  \ifdim\@tempa\p@<\p@\scalebox{\@tempa}{\usebox\pandoc@box}%
  \else\usebox{\pandoc@box}\fi%
}
\makeatother
\graphicspath{{resources/}}

%% file: sections/abstract.tex
Progress in long-horizon agents depends on scalable, high-quality interaction trajectories. However, trajectories generated by proprietary models typically lack accessible internal reasoning, while human interaction traces lack explicit chain-of-thought (CoT), limiting the reuse of existing data for agent training and iterative improvement. To unlock this underutilized supervision, we introduce CoT-Factory, a framework for training specialized CoT synthesis models that reconstruct reasoning from recorded histories and target actions without rebuilding or replaying interaction environments. Specifically, we adapt Tree-SFT to reduce redundant history computation and enable scalable training of synthesis models. Turning to the sampling stage, we introduce a new structured quality control scheme to improve reasoning quality, alongside action-conditioned guidance to improve action compatibility. On Terminal-Bench 2.1, our method improves Qwen3.5-4B to an average success rate of 26.96\%, approaching the ideal upper bound of 28.46\% achieved with original CoT. Across multiple agent benchmarks, our method consistently outperforms the compared CoT synthesis methods. These results demonstrate that CoT-Factory converts existing action trajectories into effective reasoning supervision, advancing inverse CoT synthesis for long-horizon agent training.

%% file: figures/training-progress-teaser.tex
\begin{figure}[t]
  \centering
  \includegraphics[width=\linewidth]{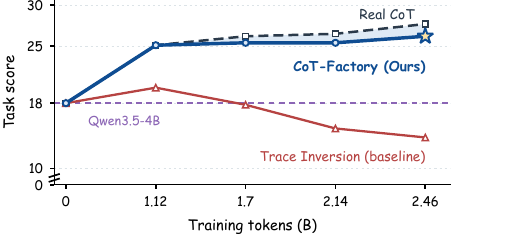}
  \caption{\textbf{Overview of CoT-Factory's performance on Terminal-Bench 2.1.} Real CoT represents the ideal performance upper bound for synthesis methods. While the baseline degrades model performance, CoT-Factory provides high-quality synthetic data that progressively improves agent capabilities with training and approaches this ideal upper bound.}
\label{fig:training-progress-teaser}
\end{figure}

%% file: sections/introduction.tex
\section{Introduction}\label{introduction}

Frontier long-horizon agents are advancing from tool use \citep{toolformer, toolllm} toward autonomous task completion \citep{swebench, osworld, terminalbench, workarena, browsergym}. Their progress increasingly depends on large-scale, high-quality interaction trajectories, collected through rollouts in newly constructed environments or curated from existing human activity records \citep{terminalhorizon, agentbank, agentgym}. Yet a major challenge in expanding these datasets is obtaining the reasoning behind recorded actions. For example, collecting successful trajectories in challenging synthetic sandboxes often requires frontier models \citep{swegym, swesmith}, whose internal chain-of-thought (CoT) is frequently withheld for safety reasons \citep{codexlogs}, while open-source models may struggle to produce successful trajectories. Meanwhile, human web interactions and code commits provide abundant action records \citep{mind2web, weblinx, commitpack}, but lack explicit CoT annotations and available environments for reconstruction. Together, these factors create a substantial imbalance between actions and high-quality CoT in public data, limiting our ability to unlock the potential of existing trajectories for agent training.

In multi-turn agents, CoT connects historical observations to the next action by identifying relevant information, diagnosing environment feedback, and weighing alternative plans \citep{react, reflexion, cot}. Action-only supervision tends to teach the surface form of an action rather than why it should be selected in the current state. Thus, \textbf{how to synthesize effective CoT from existing context and target actions without access to the original environment} is central to recovering the training value of existing trajectories, yet remains insufficiently studied for long-horizon agents.

Existing CoT completion approaches fall into two categories, both with limitations in long-horizon settings. \textbf{Forward rejection sampling} asks an open-source model to regenerate reasoning and an action from the same history, retaining only samples that match the recorded action \citep{star, restem}. Matching complex tool calls or code operations is constrained by the open-source model's capabilities, requiring repeated sampling and leaving many steps uncompleted. \textbf{Action-conditioned inverse CoT synthesis} follows two routes. Prompt-based methods directly instruct an LLM to generate CoT \citep{opencua, webstar}, but often produce limited diversity and overly simple reasoning. Methods such as RLT \citep{rlt} and Trace Inversion \citep{traceinversion} instead train specialized CoT synthesis models to generate reasoning conditioned on target outputs. However, applying these models to long-horizon agent trajectories introduces three practical challenges: low compatibility between generated reasoning and target actions; prohibitive training costs from repeatedly encoding histories of hundreds of thousands of tokens; and unreliable quality assessment, because action likelihood measures how predictable an action is without identifying factual errors or logical inconsistencies in the generated reasoning.

In this work, we introduce \textbf{CoT-Factory}, a framework that turns existing action trajectories into reasoning-augmented data for agent training without reconstructing their environments. CoT-Factory first trains a synthesis model to infer CoT from recorded context and target actions, then uses this model to complete missing CoT in existing trajectories for downstream student training. To make synthesis-model training efficient on long context, we collect teacher-generated CoT--action pairs at recorded steps and use Tree-SFT to share repeated history computation \citep{treetraining, arealdta}. During CoT synthesis, action-conditioned classifier-free guidance (CFG) improves compatibility with given actions, while structured quality control checks the generated CoT against historical evidence. The resulting trajectories provide reasoning supervision for training the student agent.

Experiments demonstrate that CoT-Factory improves efficiency and reduces costs during training and sampling compared with existing synthesis and rejection-sampling methods. Moreover, evaluations across multiple widely-used agent benchmarks show that data produced by CoT-Factory yields substantially stronger downstream performance than other synthesis baselines, approaching the ideal upper bound on several metrics and recovering an average of 68\% of the performance gains from original-CoT data across all benchmarks.

Our main contributions are as follows:
\begin{itemize}[leftmargin=*,topsep=3pt,itemsep=2pt,parsep=0pt]
\item We systematically study reasoning completion for long-horizon action trajectories without CoT, analyze the limitations of forward sampling and existing CoT synthesis methods, and validate a replay-free completion pipeline through downstream agent training experiments.
\item We adapt Tree-SFT to action-conditioned CoT synthesis, reducing training costs for more than 70,000 steps with paths up to 262,144 tokens.
\item We combine history-grounded structured quality control with action-conditioned CFG to increase candidate acceptance while reducing synthesis cost.
\item We establish a practical recipe for large-scale, replay-free reasoning completion, showing how synthesis-training scale, structured verification, and reasoning diversity jointly support effective downstream supervision.
\end{itemize}

\input{figures/method-overview.tex}

%% file: figures/method-overview.tex
\begin{figure*}[t]
\centering
\includegraphics[width=.90\textwidth]{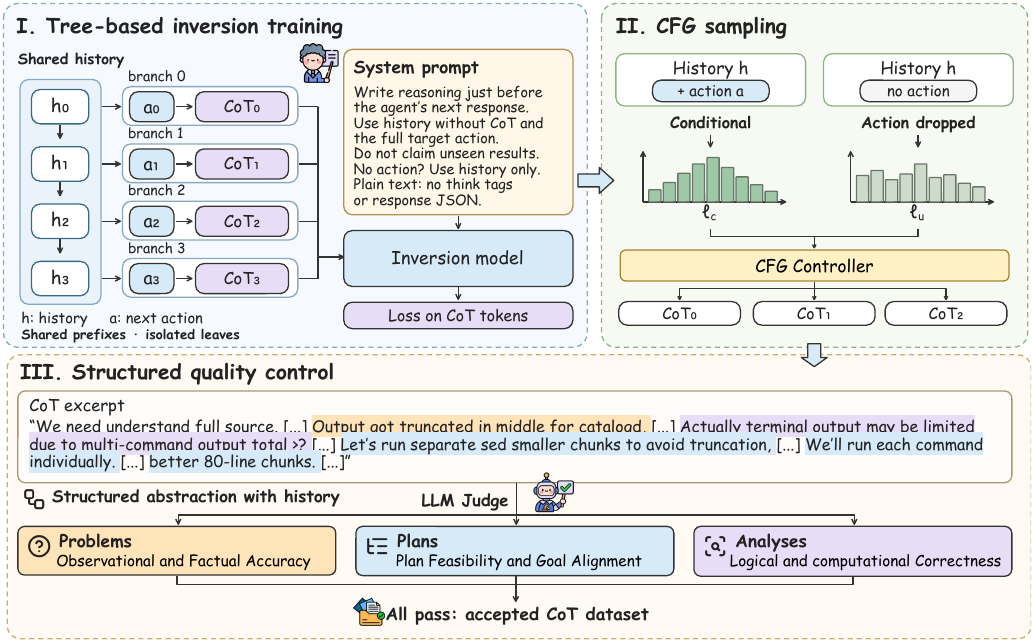}
\caption{\textbf{CoT-Factory overview.} Tree-SFT learns the CoT synthesis model, action-conditioned CFG improves sampling compatibility, and structured quality control verifies problems, plans, and their analyses against recorded history.}
\label{fig:method-overview}
\end{figure*}

%% file: sections/related-work.tex
\section{Related Work}\label{related-work}

\subsection{Scalable Agent Data}
\label{sec:related-agent-data}

\noindent\textbf{Synthetic data.} SWE-Gym, SWE-smith, and TerminalHorizon \citep{swegym, swesmith, terminalhorizon} scale software tasks and trajectories, while OS-Genesis and ComputerRL \citep{osgenesis, computerrl} extend GUI task generation and online RL. AgentOhana \citep{agentohana} unifies heterogeneous agent trajectories for training, while OS-ATLAS \citep{osatlas} scales GUI grounding data and action modeling. Such data can provide verifiable outcomes without accessible teacher reasoning.

\noindent\textbf{Real-world data.} Mind2Web and WebLINX \citep{mind2web, weblinx} collect human web demonstrations; Android in the Wild and AndroidControl \citep{aitw, androidcontrol} cover mobile interactions, and OctoPack \citep{commitpack} uses code commits. These records often lack decision-time CoT. We recover reasoning supervision from both synthetic and real-world action records without environment replay.

\subsection{Inverse CoT Synthesis}
\label{sec:related-inverse-reasoning}

\noindent\textbf{Prompt-based synthesis.} OpenCUA \citep{opencua} synthesizes reflective reasoning for recorded state--action pairs, while WebSTAR \citep{webstar} augments action trajectories with intermediate thoughts and applies step-level filtering.

\noindent\textbf{Trained synthesis models.} Trace Inversion, RLT, and VR \citep{traceinversion, rlt, vr} train models to recover reasoning from known outputs or student feedback. We study this direction in long-horizon agent trajectories, where reasoning must connect extensive interaction histories to recorded actions. In this setting, making the target action likely does not ensure that the reconstructed reasoning is consistent with the preceding observations.

%% file: sections/method.tex
\section{Method: CoT-Factory}\label{method-cot-factory}

CoT-Factory combines Tree-SFT, teacher-branch training data, action-conditioned CFG, and historical verification to complete reasoning without reconstructing or replaying the recorded interaction environments (Figure~\ref{fig:method-overview}).

\subsection{Preliminaries}\label{sec:preliminary}

Let \(u\) be the task instruction, \(a_t\) the recorded action at step \(t\), and \(o_t\) the observation returned after executing \(a_t\) in a \(T\)-step trajectory. The trajectory without CoT is \(\tau^-=(u,a_1,o_1,\ldots,a_T,o_T)\). Before action \(a_t\), the recorded history is \(h_t=(u,a_1,o_1,\ldots,a_{t-1},o_{t-1})\), with \(h_1=(u)\). The synthesis model \(q_\theta\), with parameters \(\theta\), generates CoT \(r_t\) conditioned on \(h_t\) and \(a_t\), without access to \(o_t\) or later observations. Each step uses only recorded information, enabling parallel completion without environment replay or previously synthesized reasoning.

Prior methods use answer likelihood to optimize teachers or learn reasoning posteriors \citep{rlt, vr}. Let \(p_\phi\) be the scoring model with parameters \(\phi\). Omitting the step index, \(a_j\) is action token \(j\), and \(a_{<j}\) its prefix. For an action of \(m\) tokens, its negative log-likelihood is
\[\ell_\phi(a\mid h,r)=-\sum_{j=1}^{m}\log p_\phi(a_j\mid h,r,a_{<j}).\]
This measures action predictability rather than factual reasoning. A meaning-preserving transformation \(\mathcal T(r)\) can change token probabilities: if each absolute log-probability ratio is bounded by \(\epsilon\), the total deviation satisfies \(|\ell_\phi(a\mid h,\mathcal T(r))-\ell_\phi(a\mid h,r)|\leq m\epsilon\). Length normalization preserves candidate rankings for a fixed action. Figure \ref{fig:action-nll-counterexamples} shows lower-NLL candidates containing arithmetic or state-analysis errors, motivating explicit checks against historical evidence alongside the likelihood assigned to the corresponding target action under each candidate explanation.

\input{figures/action-nll-counterexamples.tex}

\subsection{Tree-SFT}\label{sec:treesft}

\input{algorithms/tree-sft.tex}

Given paired data \(\mathcal D_{\mathrm{syn}}=\{(h,a,r)\}\), \(r_j\) denotes reasoning token \(j\) and \(r_{<j}\) its prefix. The loss \(\mathcal L_{\mathrm{syn}}\) supervises only these tokens, including termination, with \(N\) denoting the batch's supervised-token count:
\[\mathcal L_{\mathrm{syn}}=-\frac{1}{N}\sum_{(h,a,r)\in\mathcal D_{\mathrm{syn}}}\sum_j\log q_\theta(r_j\mid h,a,r_{<j}),\]
Writing \(H_i\) for a history segment and \(\oplus\) for concatenation, independent expansion repeats growing histories: prefixes \(H_1\), \(H_1\oplus H_2\), and \(H_1\oplus H_2\oplus H_3\) each encode the same initial prefix \(H_1\), recomputing identical history separately for each of the three supervised examples.

Inspired by Tree Training \citep{treetraining} and AReaL-DTA \citep{arealdta}, we apply shared-prefix training to action-conditioned CoT synthesis on long-horizon agent trajectories. We refer to this application as Tree-SFT. Tree-SFT encodes shared history as one trunk and attaches action-conditioned reasoning leaves at each step. A leaf reads only the preceding trunk and its own causal prefix; positions continue from the attachment depth, preserving the same conditioning as independently processed training examples.

\input{figures/prefix-sharing.tex}

For hybrid architectures \citep{gateddelta}, attention layers reuse trunk KV, recurrent layers inherit attachment-state snapshots, and local convolutions copy their required context windows. Backpropagation scatter-adds leaf contributions to trunk KV, states, and convolution context before completing the trunk backward pass. Leaves are processed in bounded chunks to control memory (Algorithm \ref{alg:tree-training}).

With prefix length \(p_t\) and leaf length \(\ell_t\) in tokens, independent expansion and Tree-SFT process \(L_{\mathrm{linear}}\) and \(L_{\mathrm{tree}}\) tokens:
\[\begin{aligned}L_{\mathrm{linear}}&=\sum_{t=1}^{T}(p_t+\ell_t),\\L_{\mathrm{tree}}&=p_T+\sum_{t=1}^{T}\ell_t.\end{aligned}\]
Prefix reuse accelerates inference \citep{sglang, hydragen}; Tree-SFT additionally preserves training gradients and all supervised leaves. Token counts include repeated split boundaries, and measured runtime includes attention computation and scheduling costs for the supervised reasoning branches.

In Algorithm~\ref{alg:tree-training}, branch \(b\) attaches at \(t_b\) with action \(a_b\) and target \(r_b\). Shared states \(C\) are restricted to that prefix by \(C_{\preceq t_b}\). Accumulated gradients \(g_\theta\) and \(g_C\) update parameters and trunk states: leaf gradients are summed before trunk backpropagation, preserving the combined loss gradient and each branch's causal context.

\subsection{Teacher Branches Along Recorded Trajectories}\label{sec:branch-sampling}

Training a CoT synthesis model first requires high-quality history--action--reasoning pairs. Requiring the teacher to reproduce the recorded action \(a_t\) would reduce this process to costly forward rejection sampling. Instead, at each history prefix \(h_t\) in an auxiliary trajectory subset, an open-source teacher policy \(\pi_{\mathrm{tea}}\) samples reasoning \(\tilde r_t\) and action \(\tilde a_t\) in \textbf{one forward branch sample}:
\[(\tilde r_t,\tilde a_t)\sim\pi_{\mathrm{tea}}(\cdot\mid h_t).\]
Training the CoT synthesis model requires a coherent reasoning--action pair under the same history, not reproduction of \(a_t\). We filter malformed actions, empty or truncated reasoning, repetition, and action-protocol leakage. Reversing the sampled pair to \((h_t,\tilde a_t)\rightarrow\tilde r_t\) provides training data for Tree-SFT while retaining the teacher's reasoning distribution.

\input{figures/tree-sampling.tex}

Figure~\ref{fig:tree-sampling} illustrates this distinction: at step 16, the recorded action writes a release audit, whereas the teacher first inspects commit history and feature records. This alternative reasoning--action pair provides training data for the CoT synthesis model without replacing the recorded action or modifying the recorded history for subsequent steps.

\subsection{Action-Conditioned CFG Sampling}\label{sec:cfg}

For histories containing hundreds of thousands of tokens, the history prior can dilute the comparatively short target action condition, causing ordinary autoregressive samples to drift away from that action. Inspired by diffusion CFG \citep{diffusioncfg} and its autoregressive extension \citep{cfg}, we drop the current action condition in a fraction \(\rho\) of training branches. The same model thus supports both action-conditioned and action-unconditioned prediction. At reasoning token \(j\), conditional logits \(\ell_c\) and unconditional logits \(\ell_u\) are combined by extrapolation using a guidance weight \(w\geq1\), where \(\ell_{\mathrm{cfg}}\) is the guided logit vector, \(T_{\mathrm{temp}}\) the sampling temperature, and \(\hat r_{t,j}\) the sampled token:
\[\begin{aligned}\ell_{\mathrm{cfg}}&=\ell_u+w(\ell_c-\ell_u),\\\hat r_{t,j}&\sim\operatorname{Softmax}(\ell_{\mathrm{cfg}}/T_{\mathrm{temp}}).\end{aligned}\]
The difference \(\ell_c-\ell_u\) isolates the contribution of the action condition from the shared history, emphasizing decision evidence relevant to \(a_t\). For \(w>1\), CFG strengthens target-action conditioning without training a separate model and increases single-sample compatibility.

\subsection{Structured Verification of Reasoning Paths}\label{sec:qc}

Action compatibility does not ensure correct intermediate reasoning. Expressed reasoning need not faithfully reflect model computation \citep{faithfulness, biasedcot}. In our initial experiment with normal LLM-based QC on free-form CoT, whose problems, analyses, and plans are interleaved (Figure~\ref{fig:action-nll-counterexamples}), we observed inconsistent judgments across repeated evaluations of the same candidate---alternately accepting it or flagging hallucinations---leading to unstable candidate selection. Inspired by process supervision \citep{tot, lightmanprm}, Structured QC first rejects empty, truncated, repetitive, or extensively copied action text. A semantic matcher \(M(h,r,a)\) then checks whether the candidate can reasonably lead to the target action under history \(h\).

An extractor produces a set \(\mathcal S\) of \(J\) units, each with type \(z_j\), statement \(c_j\), and supporting analysis \(b_j\):
\[\mathcal S(h,r)=\{(z_j,c_j,b_j)\}_{j=1}^{J},\quad z_j\in\{\text{problem},\text{plan}\}.\]
Here \(c_j\) is the statement and \(b_j\) the explicitly expressed analysis; absent analysis is \(\varnothing\). The verifier \(V(h,z_j,c_j,b_j)\) checks historical support, factual and arithmetic consistency, and whether plans misrepresent anticipated outcomes as observations. The Boolean acceptance gate \(G\) requires
\[G(h,r,a)=M(h,r,a)\land\bigwedge_{j=1}^{J}V(h,z_j,c_j,b_j).\]
We retain the first passing candidate within a sampling budget of \(K\) attempts. Unresolved steps preserve observations and actions, but their unverified reasoning is masked from student loss. Verification uses neither reference CoT nor future observations and requires no replay. Empirical Pass@\(K\) measures coverage: Structured QC screens the reasoning candidates, while CFG improves their acceptance rate and increases the fraction of recorded interaction steps covered by accepted reasoning.

%% file: figures/action-nll-counterexamples.tex
\begin{figure*}[t]
\centering
\includegraphics[width=\textwidth]{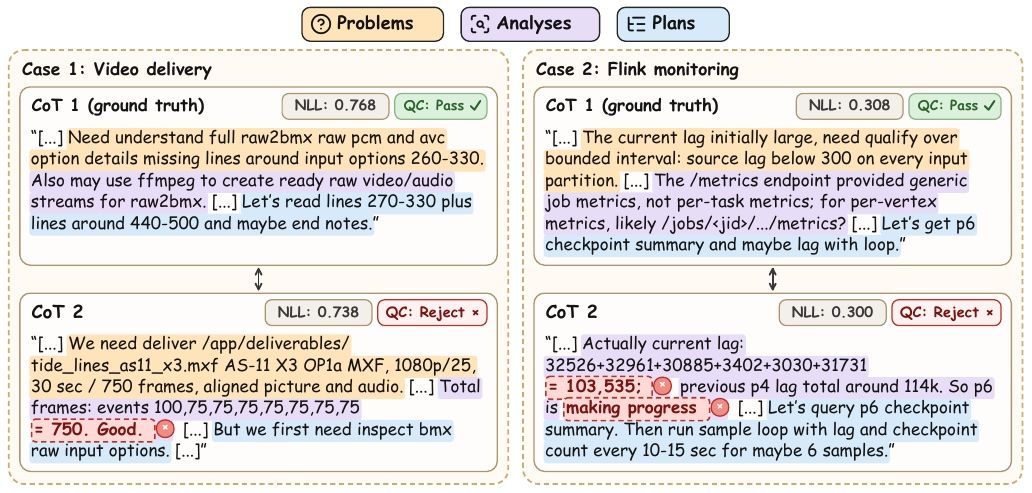}
\caption{\textbf{Lower action NLL does not guarantee correct reasoning.} In each case, CoT 2 has lower NLL than the ground-truth CoT 1 but contains arithmetic or state-analysis errors. Orange, purple, and blue indicate problems, analyses, and plans.}
\label{fig:action-nll-counterexamples}
\end{figure*}

%% file: algorithms/tree-sft.tex
\begin{algorithm}[t]
\caption{Tree-SFT forward and backward}
\label{alg:tree-training}
\small
\begin{algorithmic}[1]
\Require Histories $H$, branches $\mathcal B=\{(t_b,a_b,r_b)\}$, parameters $\theta$
\State $C\gets\operatorname{TrunkForward}_{\theta}(H)$
\State $N\gets\sum_{b\in\mathcal B}|r_b|$; $g_\theta\gets0$, $g_C\gets0$
\For{each branch $b\in\mathcal B$}
  \State $q_b\gets\operatorname{LeafForward}_{\theta}(a_b,r_b;C_{\preceq t_b})$
  \State $\mathcal L_b\gets-N^{-1}\sum_j\log q_b(r_{b,j}\mid r_{b,<j})$
  \State $(g_{\theta,b},g_{C,b})\gets\operatorname{LeafBackward}(\mathcal L_b)$
  \State $g_\theta\gets g_\theta+g_{\theta,b}$
  \State $g_C\gets\operatorname{ScatterAdd}(g_C,g_{C,b},t_b)$
\EndFor
\State $g_\theta\gets g_\theta+\operatorname{TrunkBackward}(C,g_C)$
\State $\theta\gets\operatorname{OptimizerStep}(\theta,g_\theta)$
\State \Return $\theta$
\end{algorithmic}
\end{algorithm}

%% file: figures/prefix-sharing.tex
\begin{figure}[t]
\centering
\includegraphics[width=\linewidth]{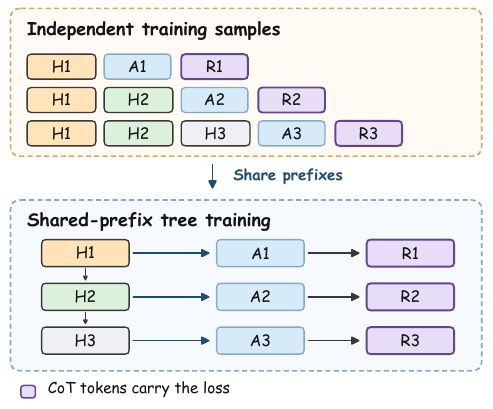}
\caption{\textbf{Illustration of how Tree-SFT reduces redundant token computation.} H denotes action--observation history accumulated along the trunk, A the action, and R the CoT target. Shared history is computed once; only CoT tokens receive direct supervision during training.}
\label{fig:prefix-sharing}
\end{figure}

%% file: figures/tree-sampling.tex
\begin{figure}[t]
\centering
\includegraphics[width=\linewidth]{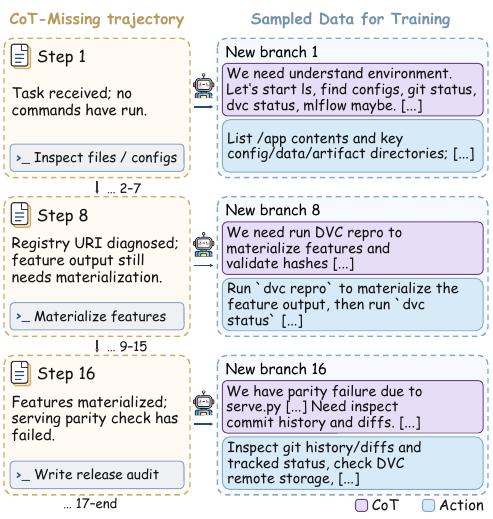}
\caption{\textbf{Tree sampling for CoT synthesis model training.} A teacher generates independent CoT--action pairs as leaves from trunk histories containing actions and observations, but no CoT. These pairs need not match the ground truth; they train a CoT synthesis model to generate CoT conditioned on history and action.}
\label{fig:tree-sampling}
\end{figure}

%% file: sections/experiments.tex
\section{Experiments}\label{experiments}

\input{tables/main-results.tex}

\subsection{Settings}\label{setup-and-downstream-performance}

\noindent\textbf{Data and training.} We reconstruct reasoning for TerminalHorizon-3k trajectories \citep{terminalhorizon}, withholding original CoT from reconstruction and retaining it for the Real and mask-CoT references. We sample reasoning--action pairs with DeepSeek-V4.1-Flash at 70,724 valid steps from 1,341 auxiliary trajectories to train the CoT synthesis model. We use Qwen3.5-4B \citep{qwen35} for CoT synthesis and evaluate Qwen3.5-4B and 9B students. Synthesis uses guidance weight \(w=2\), temperature \(T_{\mathrm{temp}}=1\), and Qwen3.8-27B for structured quality control, allowing at most eight generation attempts to complete reasoning at each recorded step.

\noindent\textbf{Benchmarks.} We report avg@3 and pass@3 on Terminal-Bench 1.0 and 2.1 (TB-1.0/2.1) \citep{terminalbench}, and success rates on GAIA \citep{gaia}, VITA-Cross \citep{vitabench}, AutomationBench \citep{automationbench}, and DeepPlanning \citep{deepplanning}. Following \citet{konakona2026}, Terminal-Bench uses six times the default agent time limits. To reduce ablation costs, TB-subset selects 50 TB-2.1 tasks whose time limits under this setting are at most 1.5 hours. Gain recovery measures how much of the rollout performance gain from original-CoT training (Real) is recovered by synthetic CoT: \(\max(0,(s-s_{\mathrm{Base}})/(s_{\mathrm{Real}}-s_{\mathrm{Base}}))\), where \(s\) is the evaluated score and references use the same model size. We take the arithmetic mean over all eight metrics in Table~\ref{tab:main}, including both avg@3 and pass@3 for TB-1.0 and TB-2.1. Recovery is 0\% at or below Base and 100\% when matching the corresponding Real reference.

\noindent\textbf{Baselines.} Base denotes the untuned model, while Real uses original-CoT supervision. mask-CoT retains original reasoning as context but supervises only action tokens. We reproduce Trace Inversion \citep{traceinversion} and evaluate RLT \citep{rlt} and VR \citep{vr} under the same SFT GPU-hour budget as our synthesis model. RLT additionally optimizes the synthesis teacher through RL using student target-action log-likelihood, while VR selects candidates by action likelihood. Trace Inversion and VR use the same eight-candidate budget per step for filtering and selection.

\subsection{Quantitative Analysis}\label{qualitative-analysis}

Preliminary no-CoT training, supervising actions with reasoning removed, yields only 3.33\% TB-2.1 avg@3---an expected catastrophic collapse motivating reasoning supervision. We therefore omit other benchmarks for this setting. Retaining reasoning as context mitigates this collapse, but mask-CoT still lowers TB-1.0 avg@3 from 28.75\% to 10.00\% (Table~\ref{tab:main}).

The quality of reconstructed reasoning is equally consequential. Under the same synthesis budget, Trace Inversion's lower acceptance rate leaves trajectories incompletely reconstructed (Table~\ref{tab:quality}), and even training on its accepted CoT underperforms Base across all benchmarks. VR-style likelihood filtering likewise fails to improve downstream performance, suggesting that action likelihood alone is insufficient to identify useful reasoning supervision. For RLT, rewards increase during continued RL while reasoning entropy and instruction following deteriorate, leaving the student at only 10.48\% avg@3 on TB-2.1.

In contrast, large-scale synthesis training and structured quality control produce useful reasoning supervision. CoT-Factory-4B improves TB-2.1 avg@3 by 8.61 points and matches Real's 40.44\% pass@3, while exceeding or approaching Real on VITA-Cross and AutomationBench, respectively. These gains extend to the 9B student, which also improves TB-2.1 avg@3 by 8.61 points and gains on both general-agent benchmarks. Across all eight metrics, mean gain recovery reaches 68.12\% and 64.24\% for the 4B and 9B students, demonstrating the value of synthesized reasoning at both model scales.

GAIA exhibits an unexpected decline at both scales. Inspection identifies an intrinsic bias in the synthesized data: under the GAIA harness, models can fail to submit a final answer. Addressing this mismatch between synthesized interaction patterns and benchmark-specific answer submission is a direction for future work.

\subsection{Candidate Acceptance, Cost, and Component Ablations}\label{candidate-acceptance-cost-and-component-ablations}

We compare single-attempt acceptance (Pass@1), cumulative acceptance within eight attempts (Pass@8), and GPU hours required to complete the full dataset under the same quality criteria (Table \ref{tab:quality}). Costs include failed candidates and retries. VR is excluded because it is a posterior selection mechanism.

\input{tables/candidate-acceptance.tex}

Forward rejection sampling must find matching actions in a large action space, achieving only 40.74\% Pass@1 and leaving 26.27\% of steps unresolved after eight attempts, at a cost of 1,243 GPU hours. RLT's hallucinated reasoning lowers single-attempt acceptance to 35.52\%, costing 1,391 GPU hours. Our synthesis model trained on 70k steps achieves 69.54\% Pass@1 and reduces synthesis cost to 697 GPU hours. Adding action-conditioned CFG raises Pass@1 by almost 20 points to 89.36\% and Pass@8 to 99.84\%. The high acceptance rate reduces repeated failures enough to lower total GPU time further, even after accounting for CFG's decoding overhead.

\input{figures/cfg-cost-sft-results.tex}

\noindent\textbf{Robustness under long contexts.} Figure \ref{fig:cfg-cost-sft-results}(a) groups 6,619 validation steps by context length. Without CFG, Pass@1 falls from 85.4\% for histories shorter than 16k tokens to 63.3\% at 64--128k tokens as the action condition is diluted. CFG with \(w=2\) improves all five length groups, including a 23.9-point gain at 64--128k tokens, and maintains similar acceptance across several groups. These results support stronger action conditioning and instruction following when reasoning is generated from long interaction histories.

\noindent\textbf{Training scale and verification.}
Table~\ref{tab:ablation} evaluates synthesis-training scale, verification, and action-conditioned CFG on TB-subset. The first two rows use the first format-valid candidate and compare 10k with 70k synthesis-training steps. Tree-SFT preserves the per-example training objective through shared-prefix computation; this comparison evaluates the larger training scale it enables, while its computational benefit is evaluated separately at matched training scales in Section~\ref{tree-structured-training-efficiency}. Normal QC directly asks an LLM to assess free-form CoT, whereas Structured QC extracts and verifies reasoning units. These are alternative filters applied to the same 70k synthesis model; the final row adds CFG with guidance weight \(w=2\) to Structured QC.

\input{tables/component-ablation.tex}

Scaling synthesis training from 10k to 70k steps raises downstream avg@3 from 23.3\% to 31.3\%. Normal QC provides no improvement, whereas Structured QC increases avg@3 to 37.3\%, supporting the value of structured verification. Consistent with this difference, qualitative checks reveal that direct LLM evaluations of the same CoT alternate between acceptance and hallucination flags, making candidate selection unstable. Adding CFG primarily improves synthesis efficiency, as shown above, while maintaining comparable downstream performance at 38.0\% avg@3.

\subsection{Tree-SFT Efficiency}\label{tree-structured-training-efficiency}

We compare independently expanded training with Tree-SFT on the training set of 1,341 long trajectories and 70,724 synthesis steps. Root-to-leaf paths reach 262,144 tokens; training runs for three epochs on 16 NVIDIA B200 GPUs. Accounting retains all 70,724 supervised leaves and includes repeated boundary prefixes required to split trajectories that exceed the maximum supported context length.

Independent expansion processes 4.576B tokens per epoch, whereas trunk reuse reduces this to 397.70M, an observed \(11.51\times\) compression. Over the full three-epoch training run, Tree-SFT requires 85.89 GPU hours, compared with 471 GPU hours for independent expansion without prefix sharing (Figure \ref{fig:cfg-cost-sft-results}(b)). Across four supervision scales from 10k to 70k steps, the savings remain between \(4.65\times\) and \(5.49\times\), substantially reducing the computational bottleneck in long-horizon inverse CoT synthesis.

\input{tables/tree-training-efficiency.tex}

Table \ref{tab:efficiency} groups the trajectories by valid synthesis-step count, illustrating the reduction of historical redundancy from quadratic to linear growth. Longer trajectories contain more shared prefixes and benefit more from Tree-SFT. As trajectory length increases from 11--20 to over 60 steps, independently expanded token volume increases from 412.1K to 9.45M per trajectory, whereas Tree-SFT requires only 685.7K in the longest group. Compression rises from \(4.65\times\) to \(13.77\times\), and the calibrated speed ratio increases from \(2.22\times\) to \(6.57\times\). Trajectories longer than 20 steps constitute 92.3\% of this TH-3k-derived set, highlighting the importance of Tree-SFT for reasoning synthesis models trained on long-horizon interaction trajectories.

\subsection{Student Loss and Sampling Temperature}\label{student-loss-and-sampling-temperature}

\noindent\textbf{Diversity collapse and temperature ablation.} We observe that the diversity of synthesized CoT is closely related to student-loss dynamics and downstream generalization. Figure \ref{fig:cfg-cost-sft-results}(c) compares students trained on Real CoT, CoT-Factory, and RLT data over approximately 2.4B tokens. RLT's strong optimization toward single-step action likelihood is associated with diversity collapse: its initial loss is approximately 0.95, unusually below that of real data. We also investigate sampling temperature \(T_{\mathrm{temp}}\in\{0,0.3,0.5,0.7,1.0\}\) to assess reasoning diversity (Table \ref{tab:temperature}). Under greedy decoding, resampling cannot escape deterministic local errors, and QC Pass@8 is only 67.3\%. Higher temperatures progressively improve downstream performance. From \(T_{\mathrm{temp}}=0.3\) onward, candidates at all tested temperatures largely pass quality control, yet increasing temperature yields 6.67 additional points on TB-subset avg@3. We interpret this as evidence that the synthesis model captures multiple valid reasoning paths: low temperatures concentrate on a few paths, whereas higher temperatures sample a broader range of correct explanations, providing richer supervision for complex trajectories, consistent with the value of diverse reasoning paths \citep{selfconsistency}.

\input{tables/temperature-ablation.tex}

%% file: tables/main-results.tex
% Editable master table. Fill TB-1.0 cells directly; build preserves this file.
\begin{table*}[t]
\centering
\begingroup\setlength{\fboxsep}{1pt}
\caption{\textbf{Downstream performance (\%).} References and experimental methods are separated; \colorbox{blue!2}{\textbf{blue}} and \colorbox{orange!4}{\textbf{orange}} name cells denote 4B and 9B models, respectively. Superscripts show percentage-point changes from the corresponding Base. Gain recovery summarizes the fraction of Real-over-Base improvement recovered averaged equally over all eight reported metrics.}
\label{tab:main}
\endgroup
\begingroup
\fontsize{8}{9.6}\selectfont\renewcommand{\tabularxcolumn}[1]{m{#1}}
\setlength{\tabcolsep}{2pt}
\renewcommand{\arraystretch}{1.12}
\begin{tabularx}{\linewidth}{@{}>{\raggedright\arraybackslash}m{.20\linewidth}*{9}{>{\centering\arraybackslash}X}@{}}
\toprule
\multirow{3}{*}[-2pt]{\textbf{Method}} & \multicolumn{5}{c}{\textbf{Coding Agent}} & \multicolumn{3}{c}{\textbf{General Agent}} & \multirow{3}{*}[-2pt]{\shortstack{\textbf{Gain}\\\textbf{recovery}}} \\
\cmidrule(lr){2-6}\cmidrule(lr){7-9}
 & \textbf{TB-1.0} & \textbf{TB-1.0} & \textbf{TB-2.1} & \textbf{TB-2.1} & \multirow{2}{*}{\textbf{GAIA}} & \multirow{2}{*}{\shortstack{\textbf{VITA-}\\\textbf{Cross}}} & \multirow{2}{*}{\shortstack{\textbf{Automation}\\\textbf{Bench}}} & \multirow{2}{*}{\shortstack{\textbf{Deep}\\\textbf{Planning}}} & \\
 & avg@3 & pass@3 & avg@3 & pass@3 & & & & & \\
\midrule
\multicolumn{10}{l}{\textbf{References}} \\
\cellcolor{blue!4}Qwen3.5-4B (Base) \cite{qwen35} & 28.75 & 41.25 & 18.35 & 31.46 & 44.85 & 10.00 & 19.71 & 21.25 & 0.00 \\
\cellcolor{orange!8}Qwen3.5-9B (Base) \cite{qwen35} & 35.00 & 47.50 & 22.10 & 34.83 & 49.30 & 14.50 & 28.65 & 21.46 & 0.00 \\
% \cellcolor{blue!4}no-CoT \cite{terminalhorizon}& -- & -- & 3.33\textsuperscript{\fontsize{5.5}{6}\selectfont -15.02} & 4.66\textsuperscript{\fontsize{5.5}{6}\selectfont -26.80} & -- & -- & -- & -- & -- \\
\cellcolor{blue!4}mask-CoT \cite{terminalhorizon} & 10.00\textsuperscript{\fontsize{5.5}{6}\selectfont -18.75} & 16.30\textsuperscript{\fontsize{5.5}{6}\selectfont -24.95} & 19.48\textsuperscript{\fontsize{5.5}{6}\selectfont +1.13} & 33.71\textsuperscript{\fontsize{5.5}{6}\selectfont +2.25} & 43.64\textsuperscript{\fontsize{5.5}{6}\selectfont -1.21} & 10.25\textsuperscript{\fontsize{5.5}{6}\selectfont +0.25} & 21.15\textsuperscript{\fontsize{5.5}{6}\selectfont +1.44} & 24.59\textsuperscript{\fontsize{5.5}{6}\selectfont +3.34} & 12.59 \\
\cellcolor{blue!4}Real (4B) \cite{terminalhorizon} & 36.67\textsuperscript{\fontsize{5.5}{6}\selectfont +7.92} & 48.75\textsuperscript{\fontsize{5.5}{6}\selectfont +7.50} & 28.46\textsuperscript{\fontsize{5.5}{6}\selectfont +10.11} & 40.44\textsuperscript{\fontsize{5.5}{6}\selectfont +8.98} & 47.88\textsuperscript{\fontsize{5.5}{6}\selectfont +3.03} & 18.50\textsuperscript{\fontsize{5.5}{6}\selectfont +8.50} & 27.61\textsuperscript{\fontsize{5.5}{6}\selectfont +7.90} & 28.96\textsuperscript{\fontsize{5.5}{6}\selectfont +7.71} & 100.00 \\
\cellcolor{orange!8}Real (9B) \cite{terminalhorizon} & 40.00\textsuperscript{\fontsize{5.5}{6}\selectfont +5.00} & 51.25\textsuperscript{\fontsize{5.5}{6}\selectfont +3.75} & 34.08\textsuperscript{\fontsize{5.5}{6}\selectfont +11.98} & 50.56\textsuperscript{\fontsize{5.5}{6}\selectfont +15.73} & 53.33\textsuperscript{\fontsize{5.5}{6}\selectfont +4.03} & 21.50\textsuperscript{\fontsize{5.5}{6}\selectfont +7.00} & 41.46\textsuperscript{\fontsize{5.5}{6}\selectfont +12.81} & 27.71\textsuperscript{\fontsize{5.5}{6}\selectfont +6.25} & 100.00 \\
\midrule
\multicolumn{10}{l}{\textbf{Experimental methods}} \\
\cellcolor{blue!4}Trace Inversion \cite{traceinversion} & 27.08\textsuperscript{\fontsize{5.5}{6}\selectfont -1.67} & 38.75\textsuperscript{\fontsize{5.5}{6}\selectfont -2.50} & 13.48\textsuperscript{\fontsize{5.5}{6}\selectfont -4.87} & 24.71\textsuperscript{\fontsize{5.5}{6}\selectfont -6.75} & 35.15\textsuperscript{\fontsize{5.5}{6}\selectfont -9.70} & 8.75\textsuperscript{\fontsize{5.5}{6}\selectfont -1.25} & 17.10\textsuperscript{\fontsize{5.5}{6}\selectfont -2.61} & 5.41\textsuperscript{\fontsize{5.5}{6}\selectfont -15.84} & 0.00 \\
\cellcolor{blue!4}RLT \cite{rlt} & 22.92\textsuperscript{\fontsize{5.5}{6}\selectfont -5.83} & 37.50\textsuperscript{\fontsize{5.5}{6}\selectfont -3.75} & 10.48\textsuperscript{\fontsize{5.5}{6}\selectfont -7.87} & 15.73\textsuperscript{\fontsize{5.5}{6}\selectfont -15.73} & 30.30\textsuperscript{\fontsize{5.5}{6}\selectfont -14.55} & 4.50\textsuperscript{\fontsize{5.5}{6}\selectfont -5.50} & 8.61\textsuperscript{\fontsize{5.5}{6}\selectfont -11.10} & 13.33\textsuperscript{\fontsize{5.5}{6}\selectfont -7.92} & 0.00 \\
\cellcolor{blue!4}VR \cite{vr} & 25.83\textsuperscript{\fontsize{5.5}{6}\selectfont -2.92} & 37.50\textsuperscript{\fontsize{5.5}{6}\selectfont -3.75} & 13.10\textsuperscript{\fontsize{5.5}{6}\selectfont -5.25} & 26.96\textsuperscript{\fontsize{5.5}{6}\selectfont -4.50} & 33.93\textsuperscript{\fontsize{5.5}{6}\selectfont -10.92} & 9.50\textsuperscript{\fontsize{5.5}{6}\selectfont -0.50} & 18.86\textsuperscript{\fontsize{5.5}{6}\selectfont -0.85} & 5.83\textsuperscript{\fontsize{5.5}{6}\selectfont -15.42} & 0.00 \\
\cellcolor{blue!4}\textbf{CoT-Factory-4B (Ours)} & 34.58\textsuperscript{\fontsize{5.5}{6}\selectfont +5.83} & 45.00\textsuperscript{\fontsize{5.5}{6}\selectfont +3.75} & 26.96\textsuperscript{\fontsize{5.5}{6}\selectfont +8.61} & 40.44\textsuperscript{\fontsize{5.5}{6}\selectfont +8.98} & 43.03\textsuperscript{\fontsize{5.5}{6}\selectfont -1.82} & 19.00\textsuperscript{\fontsize{5.5}{6}\selectfont +9.00} & 27.23\textsuperscript{\fontsize{5.5}{6}\selectfont +7.52} & 23.96\textsuperscript{\fontsize{5.5}{6}\selectfont +2.71} & \textbf{68.12} \\
\cellcolor{orange!8}\textbf{CoT-Factory-9B (Ours)} & 38.33\textsuperscript{\fontsize{5.5}{6}\selectfont +3.33} & 50.00\textsuperscript{\fontsize{5.5}{6}\selectfont +2.50} & 30.71\textsuperscript{\fontsize{5.5}{6}\selectfont +8.61} & 48.31\textsuperscript{\fontsize{5.5}{6}\selectfont +13.48} & 45.42\textsuperscript{\fontsize{5.5}{6}\selectfont -3.88} & 20.25\textsuperscript{\fontsize{5.5}{6}\selectfont +5.75} & 36.46\textsuperscript{\fontsize{5.5}{6}\selectfont +7.81} & 26.46\textsuperscript{\fontsize{5.5}{6}\selectfont +5.00} & \textbf{64.24} \\
\bottomrule
\end{tabularx}
\endgroup
\end{table*}

%% file: tables/candidate-acceptance.tex
\begin{table}[t]
\centering
\caption{\textbf{Candidate acceptance (\%) and synthesis cost.} GPU hours include failed candidates and retries.}
\label{tab:quality}
\begingroup
\small
\setlength{\tabcolsep}{3pt}
\renewcommand{\arraystretch}{1.12}
\begin{tabular*}{\linewidth}{@{\extracolsep{\fill}}lccc@{}}
\toprule
\textbf{Method} & \textbf{Pass@1} & \textbf{Pass@8} & \textbf{GPU h} \\
\midrule
Forward & 40.74 & 73.73 & 1243 \\
Trace Inversion & 58.79 & 82.53 & 891 \\
RLT & 35.52 & 65.97 & 1391 \\
Tree-SFT & 69.54 & 88.18 & 697 \\
\textbf{Tree-SFT + CFG} & \textbf{89.36} & \textbf{99.84} & \textbf{623} \\
\bottomrule
\end{tabular*}
\endgroup
\end{table}

%% file: figures/cfg-cost-sft-results.tex
\begin{figure*}[t]
\centering
\includegraphics[width=.97\textwidth]{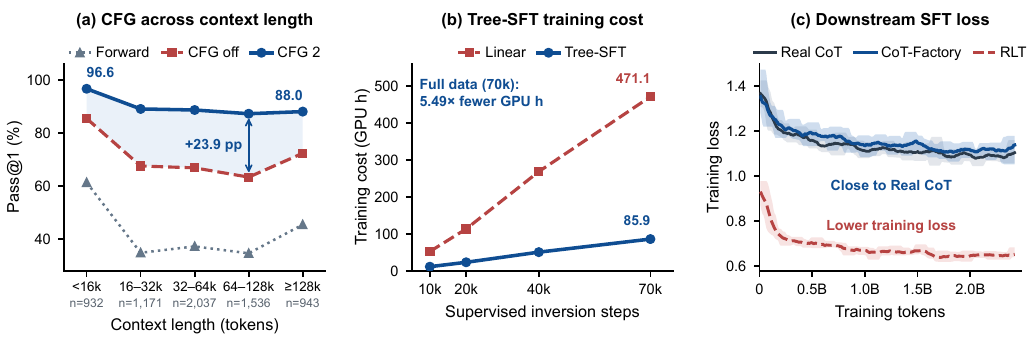}
\caption{\textbf{CFG, training efficiency, and student learning.} (a) Candidate acceptance across context lengths. (b) Training cost across supervision scales. (c) Rolling mean student loss with local min--max bands.}
\label{fig:cfg-cost-sft-results}
\end{figure*}

%% file: tables/component-ablation.tex
\begin{table}[t]
\centering
\caption{\textbf{Training scale and verification ablations on TB-subset (\%).} Normal QC and Structured QC are alternative filters applied to the 70k synthesis model; CFG is added to Structured QC with guidance weight 2.}
\label{tab:ablation}
\begingroup
\small
\setlength{\tabcolsep}{3pt}
\renewcommand{\arraystretch}{1.12}
\begin{tabular*}{\linewidth}{@{\extracolsep{\fill}}lcc@{}}
\toprule
\textbf{Configuration} & \textbf{avg@3} & \textbf{pass@3} \\
\midrule
SFT-10k & 23.3 & 38.0 \\
+ SFT-70k & 31.3 & 46.0 \\
+ Normal QC & 30.6 & 46.0 \\
+ Structured QC & 37.3 & 54.0 \\
\textbf{+ CFG} & \textbf{38.0} & \textbf{58.0} \\
\bottomrule
\end{tabular*}
\endgroup
\end{table}

%% file: tables/tree-training-efficiency.tex
\begin{table}[t]
\centering
\caption{\textbf{Tree-SFT efficiency.} Linear and Tree give mean tokens per trajectory per epoch; compression is measured and speedup is calibrated from measured linear-baseline throughput at the corresponding trajectory lengths.}
\label{tab:efficiency}
\begingroup
\footnotesize
\setlength{\tabcolsep}{3pt}
\renewcommand{\arraystretch}{1.12}
\begin{tabular*}{\linewidth}{@{\extracolsep{\fill}}lccccc@{}}
\toprule
\textbf{Steps} & $n$ & \textbf{Linear} & \textbf{Tree} & \textbf{Comp.} & \textbf{Speedup} \\
\midrule
1--10 & 2 & 152.6K & 50.7K & 3.01$\times$ & 1.43$\times$ \\
11--20 & 101 & 412.1K & 88.6K & 4.65$\times$ & 2.22$\times$ \\
21--40 & 584 & 1.07M & 148.1K & 7.25$\times$ & 3.46$\times$ \\
41--60 & 318 & 2.31M & 225.7K & 10.21$\times$ & 4.87$\times$ \\
$>60$ & 336 & 9.45M & 685.7K & 13.77$\times$ & 6.57$\times$ \\
All & 1,341 & 3.41M & 296.6K & 11.51$\times$ & 5.49$\times$ \\
\bottomrule
\end{tabular*}
\endgroup
\end{table}

%% file: tables/temperature-ablation.tex
\begin{table}[t]
\centering
\caption{\textbf{Temperature ablation on TB-subset (\%).} CoT sampling temperature strongly affects downstream performance through reasoning diversity, while QC acceptance remains nearly unchanged except under greedy decoding.}
\label{tab:temperature}
\begingroup
\small
\setlength{\tabcolsep}{3pt}
\renewcommand{\arraystretch}{1.12}
\begin{tabular*}{\linewidth}{@{\extracolsep{\fill}}cccc@{}}
\toprule
\textbf{Temperature} & \textbf{avg@3} & \textbf{pass@3} & \textbf{QC Pass@8} \\
\midrule
0.0 & 18.7 & 36.0 & 67.3 \\
0.3 & 31.3 & 50.0 & 99.5 \\
0.5 & 32.7 & 52.0 & 98.7 \\
0.7 & 36.7 & 54.0 & 99.2 \\
1.0 & 38.0 & 58.0 & 99.8 \\
\bottomrule
\end{tabular*}
\endgroup
\end{table}

%% file: sections/conclusion.tex
\section{Conclusion}\label{conclusion}

We present CoT-Factory, a framework for recovering learnable reasoning supervision from long-horizon agent trajectories without environment replay. By combining Tree-SFT, structured quality control, and action-conditioned guidance, CoT-Factory enables efficient CoT synthesis model training and high-quality reasoning synthesis. On Terminal-Bench, VITA-Cross, and AutomationBench, the synthesized supervision recovers substantial gains from original CoT and outperforms the compared CoT synthesis methods across benchmarks. Component ablations and extended experiments across model scales further validate the effectiveness and feasibility of the framework. These findings provide a scalable route to converting existing action records into reasoning supervision, extending inverse CoT synthesis to long-horizon agent learning.

%% file: sections/ai-use-statement.tex
\section*{LLM Usage Statement}
We used generative AI tools to assist with English translation, language polishing, and the presentation of figures and tables. This assistance included refining wording, improving clarity, and adjusting captions and layouts.